\documentclass[11pt,a4paper]{article}
\usepackage[hyperref]{emnlp2020}
\usepackage{times}
\usepackage{graphicx}
\usepackage{tikz}
\usepackage{amsmath}
\usepackage{arydshln}
\usepackage{enumitem}
\usepackage{comment}
\usepackage{booktabs}
\usepackage{array,multirow}
\usepackage{latexsym}

\usepackage{multicol}
\usepackage{colortbl}
\usepackage{wrapfig}
\usepackage{textcomp}

\usepackage{microtype}

\aclfinalcopy 
\def\aclpaperid{2094} 

\title{BERT Knows Punta Cana is not just \textit{beautiful}, it's \textit{gorgeous}: \\ Ranking Scalar Adjectives with Contextualised Representations}

\author{Aina Gar\'i Soler \\
  Universit\'e Paris-Saclay\\
 CNRS, LIMSI \\
 91400 Orsay, France \\
  {\tt aina.gari@limsi.fr} \\\And
  Marianna Apidianaki \\
  Department of Digital Humanities \\
  University of Helsinki \\
  Helsinki, Finland \\ 
  {\tt marianna.apidianaki@helsinki.fi} \\}

\date{}

\begin{document}
\maketitle
\begin{abstract}
Adjectives like {\it pretty}, {\it beautiful} and  {\it gorgeous} describe positive properties of the nouns they modify but with different intensity. These differences are important for natural  language understanding and reasoning. 
We propose a novel BERT-based approach to intensity detection for scalar adjectives. 
We model intensity by vectors directly derived from contextualised representations and show they can successfully rank scalar adjectives. We evaluate our models both intrinsically, on gold standard datasets, and on an Indirect Question Answering task. Our results demonstrate that BERT encodes rich knowledge about the semantics of scalar adjectives,  and is able to provide better quality intensity rankings 
than static embeddings and previous models with access to dedicated 
resources. 
\end{abstract}

\section{Introduction}

Scalar adjectives describe a property of a noun at different degrees of intensity. 
Identifying the scalar relationship that exists between their meaning (for example, the increasing intensity between  {\it pretty}, {\it beautiful} and {\it gorgeous}) is 
useful for text understanding, for both  humans and automatic systems. 
It can serve to define the sentiment and subjectivity of a text, 
perform inference and textual entailment \cite{vantieletal2016,McNally2016}, build question answering  
and recommendation systems \cite{demarneffe:10}, and assist language  learners in  distinguishing  
 between semantically similar words \cite{SheinmanandTokunaga2009}. 

\begin{figure}[t]
    \centering
    \includegraphics[width=\columnwidth]{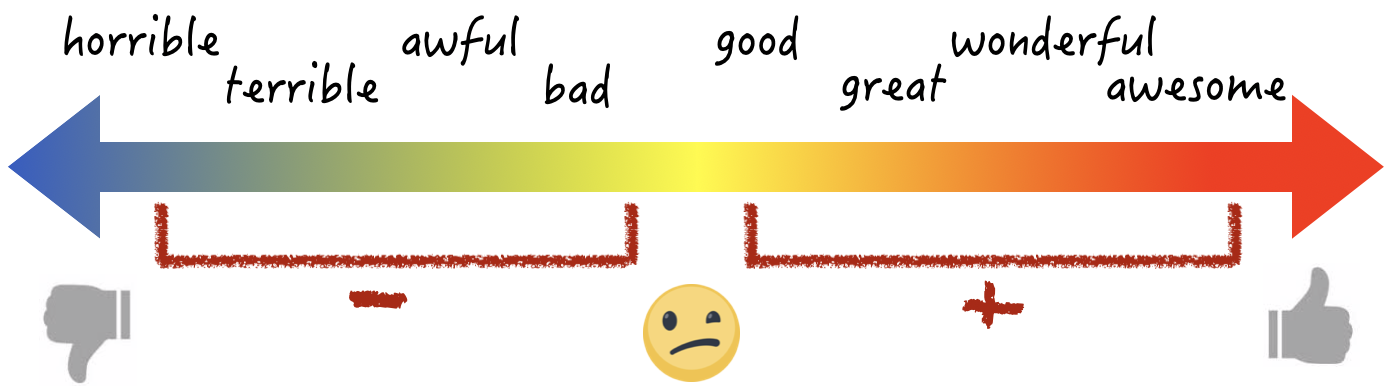}
    \caption{Full scale of adjectives describing positive and negative sentiment at different degrees from the SO-CAL dataset  \cite{taboada2011lexicon}.}
    \label{fig:elmo_plots}
\end{figure}{}

We investigate the knowledge that the pre-trained BERT  model 
\citep{devlin2019bert} encodes about the  
intensity 
expressed on an adjective scale. Given that this property is acquired by humans during language learning, we expect a language model (LM) exposed to massive amounts of text data during training to have also acquired some notion of adjective intensity. In what follows, we explore this hypothesis using representations extracted from different layers of this deep neural 
model. Since the scalar relationship between adjectives is context-dependent \cite{KennedyandMcNally2005} 
(e.g., what counts as {\it tall} may vary from context to context), 
we consider the contextualised representations produced by BERT to be a good fit for this task. We also propose a method inspired by  gender bias work
\cite{Bolukbasi-NIPS2016,devandphilips2019} for detecting the intensity relationship of two adjectives on the fly. 
We view intensity as a direction in the semantic space which, once identified, can serve to determine the intensity of new  adjectives. 

Our work falls in the neural network interpretation paradigm which explores the knowledge about language 
encoded in the representations of deep learning models 
\cite{voita-etal-2019-bottom,clark-etal-2019-bert,voita-etal-2019-analyzing,tenney-etal-2019-bert,talmor2019olmpics}. The 
bulk of this interpretation work 
addresses structural aspects of language such as syntax, word order, or number agreement \citep{linzen2016assessing,hewitt2019structural,hewitt-liang-2019-designing,rogers2020primer}; 
shallow semantic phenomena 
closely related to syntax such as semantic role labelling and coreference \cite{tenney-etal-2019-bert,kovaleva-etal-2019-revealing}; or the symbolic reasoning potential of language model representations \citep{talmor2019olmpics}. 
Our work makes a contribution towards the study of  the knowledge pre-trained LMs encode about word meaning, generally overlooked until now in interpretation work. 

We evaluate the representations generated by BERT against gold standard adjective intensity estimates \cite{demelo:13,wilkinson2017identifying,cocos-etal-2018-learning} and apply them directly to a question answering task \cite{demarneffe:10}. Our results show that BERT clearly encodes the intensity variation between adjectives on scales describing different properties. Our proposed 
method 
can be easily applied to new datasets and languages where scalar adjective resources are not available.\footnote{Our code and data are available at \url{https://github.com/ainagari/scalar_adjs}}

\section{Related Work} \label{sec:related_word}

The analysis of scalar adjective relationships in the literature has often been decomposed into two steps: Grouping related adjectives together and ranking   adjectives in the same group according to intensity. The first step can be  performed by distributional clustering approaches \citep{hatzivassiloglou:93,pang2008opinion} 
which can also address adjectival polysemy. 
{\it Hot}, for example, 
can be on the {\sc temperature} scale (a {\it warm} $\rightarrow$ {\it hot} $\rightarrow$ {\it scalding} drink), the {\sc attractiveness} (a {\it pretty} $\rightarrow$ {\it hot} $\rightarrow$ {\it sexy} person) or the {\sc interest} scale (an {\it interesting} $\rightarrow$ {\it hot} topic), depending on the attribute it modifies. 

Other works \cite{SheinmanandTokunaga2009,demelo:13,wilkinson2017identifying} directly address the second step, ranking groups of semantically related adjectives from lexicographic resources (e.g., WordNet) \cite{Fellbaum1998}. 
This ranking is the focus of this work. We show that BERT contextualised representations encode rich information about adjective intensity, and  can provide high quality rankings of adjectives in a scale. 


Adjective ranking has been traditionally performed 
using pattern-based 
approaches 
which extract lexical or syntactic  patterns indicative of an intensity relationship from large corpora \cite{SheinmanandTokunaga2009,demelo:13,sheinman2013large,shivade:15}. For example, the patterns ``{\it X, but not Y}'' and ``{\it not just X but Y}'' provide evidence that X is an adjective less intense than Y. Another common approach is lexicon-based 
and draws upon a resource that maps adjectives to  
scores encoding sentiment polarity (positive or negative) and intensity. Such resources can be manually created, like the SO-CAL lexicon \cite{taboada2011lexicon}, or 
automatically compiled by mining 
adjective orderings from star-valued product  reviews 
where people's comments have associated ratings 
\cite{demarneffe:10,Rill2012AGA,Sharma2015AdjectiveIA,Ruppenhofer2014ComparingMF}. 
\citet{cocos-etal-2018-learning} combine knowledge from lexico-syntactic patterns and the SO-CAL lexicon 
with
paraphrases in the Paraphrase Database (PPDB)  \citep{ganitkevitch2013ppdb,pavlick-etal-2015-ppdb}. 

Our approach is novel in that it does not need  specified patterns 
or access to lexicographic resources. It, instead, relies on the knowledge about intensity encoded in scalar adjectives' contextualised representations.  Our best performing method is inspired by work on gender bias which relies on simple vector arithmetic to uncover gender-related stereotypes. 
A gender direction is determined (for example, by comparing the embeddings of \textit{she} and \textit{he}, or \textit{woman} and \textit{man}) and  
the projection of the vector of a potentially biased word on this direction is then calculated  \cite{Bolukbasi-NIPS2016,zhao-etal-2018-learning}. We 
extend this method to scalar adjectives 
and BERT representations.

\newcite{kim-de-marneffe-2013-deriving} also consider vector distance in the semantic space  to 
encode scalar relationships between adjectives.
They specifically 
examine a small set of word pairs,  and observe that the middle point in space between the word2vec \cite{mikolov2013efficient} embeddings of two antonyms (e.g.,  \textit{furious} and \textit{happy}) falls close to the embedding of a mid-ranked 
word in their scale (e.g., \textit{unhappy}). Their experiments rely on antonym pairs 
extracted from WordNet. 
We show that contextualised representations are a better fit for this task than static embeddings, encoding rich information about adjectives' meaning and intensity. 

\section{Data}
\label{scalaradjdata}

We experiment with three 
scalar adjective 
datasets.

\paragraph {\sc deMelo} 
\cite{demelo:13}.\footnote{\url{http://demelo.org/gdm/intensity/}} 
Adjective sets were extracted from WordNet `dumbbell' structures \cite{gross1990adjectives}. The sets represent full-scales  
(e.g., from {\it horrible} to {\it awesome}) and are partitioned into half-scales 
(from {\it horrible} to {\it bad}, and from \textit{good} to \textit{awesome}) based on 
pattern-based evidence in the Google N-Grams corpus \cite{brants2006web}.  
The dataset contains 87 half-scales 
with 548 
adjective pairs, 
manually annotated for intensity relations ($<$, $>$, and $=$).

\vspace{2pt}
\noindent \textsc{Crowd} \cite{cocos-etal-2018-learning}.
\footnote{\url{https://github.com/acocos/scalar-adj}} The dataset consists of a set of adjective scales with high coverage of the PPDB vocabulary. It was constructed by a three-step process: Crowd workers were first asked to determine whether pairs of adjectives describe the same attribute (e.g., {\sc temperature}) and should, therefore, belong to the same scale. Sets of same-scale adjectives were then  refined over multiple rounds. Finally, workers ranked the adjectives in each set by intensity. The final dataset includes 330 
adjective pairs along 79 half-scales. 

\vspace{2pt}
\noindent \textsc{{Wilkinson}}   \cite{wilkinson2016gold}.\footnote{\url{https://github.com/Coral-Lab/scales}} This dataset was generated through crowdsourcing. Crowd workers were presented with small seed sets (e.g., \textit{huge, small, microscopic}) and were asked to propose similar adjectives, resulting in twelve adjective sets. 
Sets were automatically cleaned for consistency, and then annotated for intensity by the crowd workers. The original dataset contains full scales. 
We use its division in 21 half-scales (with 61 adjective pairs) proposed by  \newcite{cocos-etal-2018-learning}. 

In the rest of the paper, we use the term ``scale'' to refer to the half-scales contained in these datasets. 
Table \ref{fig:adjexamples} shows examples from each one of them. 

\begin{table}[t]
\begin{center}
\scalebox{0.85}{
\begin{tabular}{p{1.9cm}|p{5.5cm}}
 Dataset & Adjective scale 
 \\ \midrule
 \multirow{2}{*}{{\sc deMelo}} & [{\it soft} $\rightarrow$  {\it quiet} $\rightarrow$  {\it inaudible}  $\rightarrow$ {\it silent}] \\ 
 & [{\it thick} $\rightarrow$ {\it  dense} $\rightarrow$ {\it  impenetrable}] \\ \midrule
 
  \multirow{2}{*}{{\sc Crowd}}  & [{\it fine} $\rightarrow$ {\it  remarkable} $\rightarrow$  {\it spectacular}] \\
 & [{\it scary} $\vert\vert$ {\it frightening}  $\rightarrow$ {\it terrifying}] \\
 
 \midrule
\multirow{2}{*}{{\sc Wilkinson}} 
& [{\it damp} $\rightarrow$ {\it moist} $\rightarrow$ {\it wet}] \\ 
& [{\it dumb} $\rightarrow$ {\it stupid} $\rightarrow$ {\it idiotic}] \\ \midrule

 
\end{tabular}}
\end{center}
\caption{Examples of  
scales in each dataset. `$\vert\vert$' denotes a tie between 
adjectives of the same intensity.}
\label{fig:adjexamples}
\end{table}

\section{BERT Contextualised Representations} \label{sec:ctxt_representations}

\subsection{Sentence Collection}

To explore the knowledge BERT has about 
relationships in an adjective scale $s$, 
we generate a contextualised representation  
for each $a \in s$ in the same context. 
Since such cases are rare in running text, we construct 
two sentence sets 
that satisfy this condition using the ukWaC corpus  \cite{baroni2009wacky}\footnote{\url{http://u.cs.biu.ac.il/~nlp/resources/downloads/context2vec/}} and the Flickr 30K dataset \cite{young-etal-2014-image}.\footnote{Flickr contains 
crowdsourced captions 
for 31,783 
images describing 
everyday activities, events and scenes. 
We consider objective descriptions to be  a better fit for our task than subjective statements, 
which might contain emphatic markers.  
For example, {\it impossible} would be a  
bad substitute for {\it impractical} in the sentence ``\textit{What you ask for is  \underline{too impractical}''.}}
For every $s \in D$, a dataset 
from Section \ref{scalaradjdata}, and  for each 
$a \in s$,  we  collect 1,000 instances (sentences) 
from each corpus.
\footnote{
ukWaC has perfect coverage. Flickr 30K covers 96.56\% of the {\sc deMelo} scales and 86.08\% of the {\sc Crowd} scales. 
A scale $s$ is not covered when no $a \in s$ is found in a corpus.}  
We  substitute each  instance  
 $i$ of 
 $ a \in s$, 
 with each $ b \in s $  where $b \neq a $, 
creating 
$|s|-1$ new sentences.\footnote{We make a minor adjustment 
of the substituted data by  replacing the indefinite article $a$ with $an$ when the adjective that follows starts with a vowel, and the inverse when 
it starts with a consonant.} 
For example, for an instance of {\it thick} from the scale [{\it thick $\rightarrow$ dense $\rightarrow$  impenetrable}] in Table \ref{fig:adjexamples}, we generate two new  sentences where {\it thick} is substituted 
by each of the other  adjectives in the same context. 

\subsection{Sentence Cleaning} \label{sec:sentence_cleaning}

\paragraph{Hearst patterns} We filter out sentences where substitution should not take place, such as cases of specialisation 
or instantiation. 
In this way, we avoid  
replacing 
{\it deceptive} with {\it fraudulent} and  {\it false} in sentences like 
``{\it Viruses \underline{and other}  deceptive software}'', ``\textit{Deceptive software \underline{such as} viruses}'', ``\textit{Deceptive software, \underline{especially}  viruses}''.\footnote{This would especially be a problem when considering adjectives with different polarity on a full scale (e.g., \textit{deceptive} and \textit{honest}).} 
 We parse the sentences with {\tt stanza} \cite{qi2020stanza} to reveal their dependency structure, and use Hearst lexico-syntactic patterns  \cite{hearst-1992-automatic} to identify sentences  
  describing {\it is-a} relationships between  nouns in a text. 
  More details about this filtering 
are given in Appendix \ref{app:hearst}. 



\paragraph{Language Modelling criteria}  
Adjectives that belong to the same scale 
might not be replaceable  in all contexts. Polysemy can also influence their substitutability (e.g., {\it warm weather} is a bit {\it hot}, but a {\it warm smile} is {\it friendly}).
In order to select contexts where 
$\forall a \in s$ fit, 
we measure the fluency of the  sentences generated through substitution. We use a score assigned to each sentence by context2vec 
\cite{melamud-etal-2016-context2vec} which  
reflects how well an $a \in s$ 
fits a context by measuring the cosine similarity between $a$ and the context representation. 
We also experimented with calculating the perplexity assigned by BERT to a sentence generated through substitution, and with replacing the original $a$ instance with the [MASK] token and getting the BERT probability for each $a \in s$ as a filler for that slot. context2vec was found to make better  substitutability estimates.\footnote{We use as development set for this exploration 
a sample of 500 sentence pairs from the Concepts in Context (CoInCo) 
corpus \cite{kremer2014substitutes} that we will share along with our code. 
Details on the constitution of this sample are in Appendix \ref{app:evalsentence}.}


We use a 600-dimensional context2vec  model in our experiments, pre-trained on  ukWaC.\footnote{\url{http://u.cs.biu.ac.il/~nlp/resources/downloads/context2vec/}} We calculate the context2vec score for all  
sentences generated for a scale $s$ through substitution, and keep the ten with the lowest standard deviation ({\sc std}). 
Low {\sc std} for a sentence means that $\forall a \in s$ are reasonable choices in this context. 
For comparison, we also randomly sample ten sentences from all the ukWaC sentences collected for each scale. We call the sets of sentences ukWaC, Flickr and Random {\sc sent-set}s.

We extract the contextualised representation for each $a \in s$ in 
the ten sentences retained for scale $s$, using the pre-trained {\tt bert-base-uncased} model.\footnote{When an adjective is split into multiple wordpieces \cite{wu2016googles}, we average them to obtain its representation.} This results in $|s| * 10$ BERT representations for each scale. We repeat the procedure for every BERT layer. Examples of the obtained sentences are given in Appendix \ref{app:evalsentence}. 

\section{Scalar Adjectives Ranking} \label{sec:scalar_ranking}


\subsection{Ranking with a Reference Point} 
\label{subsec:fix_boundaries}

In our first ranking experiment, we explore whether BERT 
encodes adjective intensity relative 
to a reference point, that is the 
adjective with the highest intensity ($a_{ext}$) in a scale $s$. 

\paragraph{Method} We rank $\forall a \in s$ where $a \neq a_{ext}$ 
by intensity by measuring the cosine similarity between their   representation and that of $a_{ext}$ 
in 
the ten ukWaC sentences retained 
for $s$, and  in every BERT layer. For example, to rank 
[{\it pretty, beautiful, gorgeous}]  
we measure  
the similarity of the representations of {\it pretty} and {\it beautiful} 
to that of {\it gorgeous}. 
We then average the similarities obtained for each $a$ and use these values for ranking. We refer to this  
method as {\sc BertSim}.

\begin{table}[t]
    \centering
    \scalebox{0.9}{
    \begin{tabular}{c c | c | c c}
    Dataset & Metric & {\sc BertSim} & {\sc freq} & {\sc sense} \\
    \hline
    \parbox{1.3cm}{\multirow{3}{0pt}{{\sc deMelo}}} & {\sc p-acc} & \textbf{0.591}$_{11}$ & 0.571 & 0.493 \\
    & $\tau$ & \textbf{0.364}$_{11}$ & 0.304 & 0.192 \\
     & $\rho_{avg}$ & \textbf{0.389}$_{11}$ & 0.309 & 0.211 \\
     \hline
    \parbox{1.3cm}{\multirow{3}{0pt}{{\sc Crowd}}} & {\sc p-acc} & \textbf{0.646}$_{11}$ & 0.608 & 0.570 \\
     & $\tau$ & \textbf{0.498}$_{11}$ & 0.404 & 0.428 \\
     & $\rho_{avg}$ & 0.494$_{11}$ & 0.499 & \textbf{0.537} \\
     \hline
    \parbox{1.7cm}{\multirow{3}{0pt}{{\sc Wilkinson}}} & {\sc p-acc} & \textbf{0.913}$_{9}$ & 0.739$_{9}$ & 0.739$_{9}$ \\
    & $\tau$ & \textbf{0.826}$_{9}$ & 0.478 & 0.586 \\
    & $\rho_{avg}$ & \textbf{0.724}$_{9}$ & 0.345 & 0.493 \\

      \end{tabular}}
    \caption{{\sc BertSim} results on each dataset using contextualised representations from the ukWaC {\sc sent-set}.
    Subscripts denote the best-performing BERT layer.}
    \label{tab:oracle_bertsim_results}
\end{table}

\begin{figure}
    \centering
    \includegraphics[width=\columnwidth]{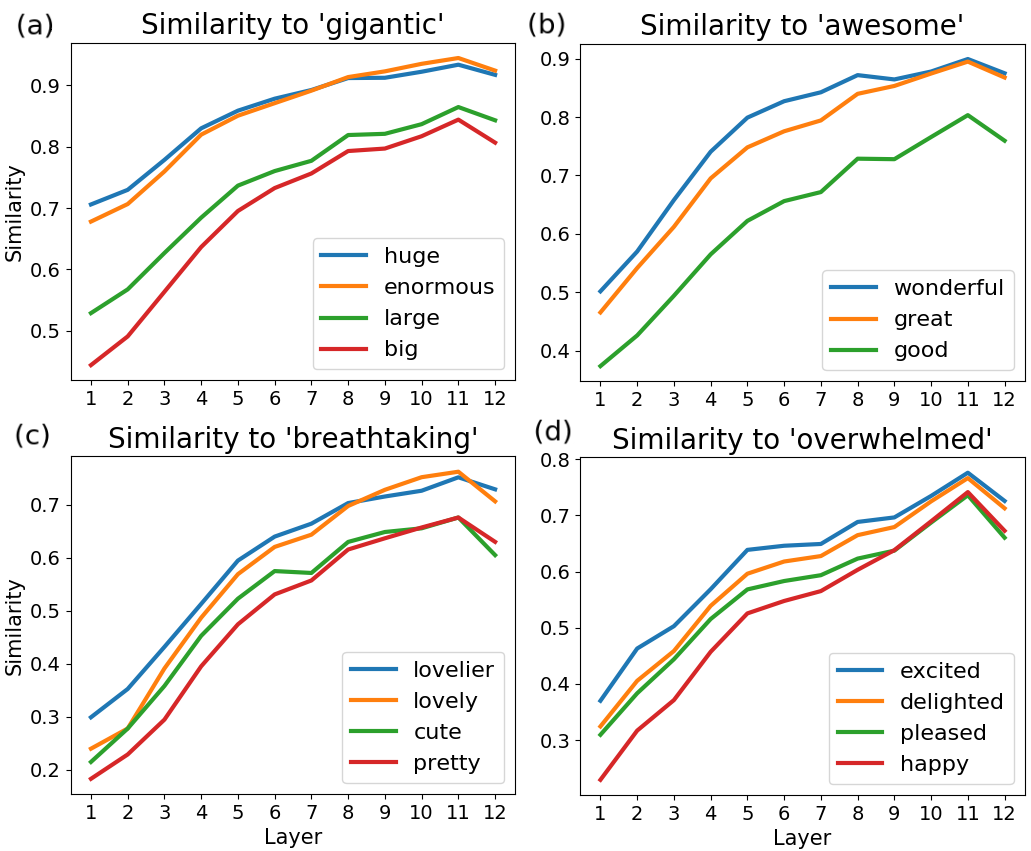}
    \caption{Examples of {\sc BertSim}  ranking predictions across layers using ukWaC sentences for four adjective scales: (a) [\textit{big} $\rightarrow$ \textit{large} $\rightarrow$ \textit{enormous} $\rightarrow$ \textit{huge}  $\rightarrow$ \textit{gigantic}], (b) [\textit{good} $\rightarrow$ \textit{great} $\rightarrow$ \textit{wonderful} $\rightarrow$ \textit{awesome}], (c) [\textit{cute} $\rightarrow$ \textit{pretty} $\rightarrow$ \textit{lovely} $\rightarrow$ \textit{lovelier} $\rightarrow$ \textit{breathtaking}], (d) [\textit{pleased} $\rightarrow$ \textit{happy} $\rightarrow$ \textit{excited} $\rightarrow$ \textit{delighted} $\rightarrow$ \textit{overwhelmed}]. (a) and (b) are from {\sc Wilkinson}, (c) and (d) are from {\sc Crowd}.}
    \label{fig:bertsim_oracle_plots}
\end{figure}

We evaluate the quality of the   
ranking for a scale 
by measuring its 
correlation with the  gold standard ranking in the corresponding dataset $D$ using Kendall's $\tau$ and Spearman's $\rho$ correlation coefficients.\footnote{
We report correlations as a weighted average using the number of adjective pairs in a scale as weights.}  We also measure the model's pairwise accuracy ({\sc p-acc})
which shows whether 
it correctly predicted the relative intensity   ($<$, $>$, $=$) for each pair $a_i$-$a_j \in s$ with $i \neq j$.  
During evaluation, 
we 
do not take into account scales where only one adjective is left ($|s| = 1$) after removing $a_{ext}$
 (26 out of 79 scales in {\sc Crowd}; 9 out of 21 scales in {\sc Wilkinson}). 
\paragraph{Baselines} We compare the {\sc BertSim} method to two baselines which rank adjectives by frequency ({\sc freq}) and  number of senses ({\sc sense}). 
    We make the assumption that 
    words with low intensity (e.g., \textit{good, old}) 
    are more frequent and  polysemous than their extreme counterparts on the same scale (e.g., \textit{awesome, ancient}).  
    This assumption relies on the following two intuitions which we empirically validate: 
    (a) Extreme adjectives tend to  restrict the denotation of a noun to a smaller class of referents than low intensity adjectives \cite{geurts2010quantity}. 
    We hypothesise that extreme adjectives denote more exceptional and less frequently encountered properties of nouns than low intensity adjectives on the same scale. This is 
    also reflected in the directionality of their entailment relationship (e.g., \textit{awesome} $\rightarrow$ \textit{good}, \textit{good} $\not\rightarrow$ \textit{awesome});  
    low intensity adjectives should thus 
    be more frequently encountered in texts. We test this assumption using frequency counts in Google Ngrams \cite{brants2006web}, 
    and find that 
   the least intense adjective is indeed more frequent than the most extreme adjective in 75\% of the scales; (b) Since frequent words tend to be more polysemous \cite{Zipf1945}, we also 
   expect that  
    low intensity adjectives would 
   have more senses than extreme ones. 
   This is confirmed 
   by their number of senses in WordNet: in 67\% of the scales, the least intense adjective has a higher number of 
   senses than its extreme counterpart.

    \paragraph{Results}
    We present the results of this evaluation in Table \ref{tab:oracle_bertsim_results}. Overall, similarities derived from BERT representations 
    encode well the notion of intensity, as shown by the  
moderate to high accuracy and correlation in the three datasets. 
The good results obtained by the {\sc freq} and {\sc sense} baselines (especially on {\sc Crowd}) 
highlight the relevance of frequency and polysemy for scalar adjective ranking, and further validate our assumptions.

Figure \ref{fig:bertsim_oracle_plots} shows 
ranking predictions made by {\sc BertSim} in different layers of the model. Predictions are generally  
stable and reasonable across layers, 
despite not always being correct. For example, the similarly-intense \textit{happy} and \textit{pleased} are inverted  
in some layers but are not confused with adjectives further up the scale (\textit{excited, delighted}). 
Note that \textit{happy} and \textit{pleased} are in adjacent positions in the {\sc Crowd} ranking, and form 
a tie 
in the {\sc deMelo} dataset. %

\subsection{Ranking without Specified Boundaries} \label{subsec:ranking_noboundaries}

In real life scenarios, scalar adjective interpretation is performed without concrete reference points (e.g., $a_{ext}$). 
We need to recognize that a {\it great book} is better than a {\it well-written} one, without necessarily detecting their relationship to {\it brilliant}. 


\begin{table*}[t]
    \centering
    \scalebox{0.9}{
     \scalebox{0.9}{
      \begin{tabular}{c  c | c | ccc | ccc | ccc}
     & & \multicolumn{1}{c}{} & \multicolumn{3}{c}{{\sc deMelo (dm)}} & \multicolumn{3}{c}{{\sc Crowd (cd)}} & \multicolumn{3}{c}{{\sc Wilkinson (wk)}} \\ \cline{1-12}
     & 
     & Method & {\sc p-acc} & $\tau$ & $\rho_{avg}$ & {\sc p-acc} & $\tau$  & $\rho_{avg}$ & {\sc p-acc} & $\tau$ & $\rho_{avg}$\\
     \hline
    \parbox[t]{2mm}{\multirow{9}{*}{\rotatebox[origin=c]{90}{BERT}}} &  
    \parbox[t]{2mm}{\multirow{3}{*}{\rotatebox[origin=c]{90}{ukWaC}}}
    & {\sc diffvec-dm} & 
     - & - & - & 
     \textbf{0.739}$_{12}$ & \textbf{0.674}$_{12}$ & \textbf{0.753}$_{12}$ &
     0.918$_{6}$& 0.836$_{6}$ & 0.839$_{6}$ \\
     & & {\sc diffvec-cd} & 
     0.646$_{8}$ & 0.431$_{8}$ & \textbf{0.509}$_{8}$ & 
     - & - & - & 
     0.869$_{11}$ & 0.738$_{11}$ & 0.829$_{11}$\\
     & & {\sc diffvec-wk} & 
     0.584$_{9}$ & 0.303$_{9}$ & 0.313$_{10}$ & 
     0.706$_{10}$ & 0.603$_{9}$ & 0.687$_{9}$ &
     - & - & - \\
    \cdashline{2-12}
    & 
    \parbox[t]{2mm}{\multirow{3}{*}{\rotatebox[origin=c]{90}{Flickr}}}
    & {\sc diffvec-dm} & 
     - & - & - & 
     0.730$_{12}$ & 0.667$_{12}$ & 0.705$_{10}$ &
     \textbf{0.934}$_{9}$ & \textbf{0.869}$_{9}$ & \textbf{0.871}$_{9}$ \\
     & &  {\sc diffvec-cd} & 
     0.620$_{10}$ & 0.377$_{10}$ & 0.466$_{10}$ & 
     - & - & - & 
     0.902$_{7}$ & 0.803$_{7}$ & 0.798$_{7}$ \\
    &  & {\sc diffvec-wk} & 
     0.579$_{1}$ & 0.294$_{1}$ & 0.321$_{1}$ & 
     0.702$_{8}$ & 0.608$_{8}$  & 0.677$_{8}$ &
     - & - & - \\
    \cdashline{2-12}
    & 
     \parbox[t]{2mm}{\multirow{3}{*}{\rotatebox[origin=c]{90}{Random}}}
    & {\sc diffvec-dm} & 
     - & - & - & 
     \textbf{0.739}$_{12}$ & 0.673$_{12}$ & 0.743$_{12}$ &
     0.918$_{6}$ & 0.836$_{6}$ & 0.839$_{6}$ \\
     & &  {\sc diffvec-cd} & 
     0.626$_{8}$ & 0.388$_{8}$ & 0.466$_{8}$ & 
     - & - & - & 
     0.836$_{12}$ & 0.672$_{12}$ & 0.790$_{10}$ \\
     & & {\sc diffvec-wk} & 
     0.557$_{9}$ & 0.246$_{9}$ & 0.284$_{6}$ & 
     0.703$_{8}$ & 0.598$_{8}$ & 0.676$_{8}$ &
     - & - & - \\
     \hline
    \parbox[t]{2mm}{\multirow{3}{*}{\rotatebox[origin=c]{90}{\small word2vec}}} & & {\sc diffvec-dm} & 
     - & - & - & 
     0.657 & 0.493 & 0.543 &
     0.787 & 0.574  & 0.663 \\
     & &  {\sc diffvec-cd} & 
     0.633 & 0.398 & 0.444 & 
     - & - & -  & 
     0.803 & 0.607  & 0.637 \\
     & & 
     {\sc diffvec-wk} & 
     0.593 & 0.323 & 0.413 & 
     0.618 & 0.413 & 0.457 &
     - & -  & - \\
     \hline
     \parbox[t]{2mm}{\multirow{3}{*}{\rotatebox[origin=c]{90}{Baseline}}}  &  & {\sc freq} &
     0.575 & 0.271 & 0.283 &
     0.606 & 0.386 & 0.452 &
     0.754 & 0.508 & 0.517 \\
     & & {\sc sense} & 
     0.493 & 0.163 & 0.165 & 
     0.658 & 0.498 & 0.595 &
     0.721 & 0.586 & 0.575 \\
     & & Cocos et al. '18 & 
     \textbf{0.653} & \textbf{0.633}  & - & 
     0.639 & 0.495 & - &
     0.754 & 0.638 & - \\
         
    \end{tabular}}}
    \caption{Results of our {\sc diffvec} 
    adjective ranking method 
    on the  {\sc deMelo}, {\sc Crowd}, and {\sc Wilkinson} datasets. 
    We report results with contextualised (BERT) representations obtained from  different {\sc sent-set}s 
    (ukWaC, Flickr, Random) and with static (word2vec) vectors.  
    We compare to the frequency ({\sc freq}) and number of senses ({\sc sense})  baselines,  
    and to results from previous work \citep{cocos-etal-2018-learning}. Results for a dataset are missing (-) when the dataset was used for building the $\overrightarrow{dVec}$ intensity vector.}
    \label{tab:diffvec_results}
\end{table*}

\paragraph{Method} Our second adjective ranking method  draws inspiration from word analogies in gender bias work, where a gender subspace 
is 
identified in word-embedding space by 
calculating 
the main direction 
spanned by the differences 
between vectors of gendered word pairs 
(e.g.,  
$\overrightarrow{he}$ - $\overrightarrow{she}$, $\overrightarrow{man}$ - $\overrightarrow{woman}$) \cite{Bolukbasi-NIPS2016,devandphilips2019,ravfogel2020null,Lauscheretal2020}. 

We propose to obtain an \textit{intensity direction} by subtracting the representation of a mild 
intensity adjective 
$a_{mild}$
from that of an extreme adjective $a_{ext}$ 
on the same scale. 
By subtracting \textit{pretty} from {\it gorgeous}, for example, which express  
a similar core meaning 
(they are both on the {\sc beauty} scale)  
but with different 
intensity, we expect the resulting $\overrightarrow{dVec}$ = $\overrightarrow{gorgeous}$ - $\overrightarrow{pretty}$ embedding to 
represent this 
notion of intensity (or degree).  
We can then compare other adjectives' representations 
to $\overrightarrow{dVec}$, and rank them according to their cosine similarity\footnote{We also tried the dot product of the vectors. The results were highly similar to the ones obtained using the cosine.} to this intensity vector: 
the closer an adjective  
is to $\overrightarrow{dVec}$, the more intense 
it is. 



We calculate the $\overrightarrow{dVec}$ for each $s \in D$ (a dataset from Section \ref{scalaradjdata}) using the most extreme ($a_{ext}$) and the mildest ($a_{mild}$) words in $s$. 
We experiment with BERT embeddings from the {\sc sent-set}s generated through substitution 
as described in Section \ref{sec:ctxt_representations}, and with  static word2vec embeddings \cite{mikolov2013efficient} trained on Google News.\footnote{We use the {\tt magnitude} library \cite{patel2018magnitude}.} 
We build a $\overrightarrow{dVec}$ from every sentence (context) $c$ in the set of ten sentences $C$ for 
a scale $s$ by subtracting the BERT representation of
$a_{mild}$ in $c$ from that of $a_{ext}$ in $c$.
We average the ten $\overrightarrow{dVec}$'s obtained for $s$ and construct a global $\overrightarrow{dVec}$ for the dataset $D$ by averaging the vectors of $\forall s \in D$. For a fair evaluation, we perform a lexical split in the data used for deriving $\overrightarrow{dVec}$ and the data used for testing.   
When evaluating on {\sc Crowd}, we calculate a $\overrightarrow{dVec}$ vector on {\sc deMelo} ({\sc diffvec-dm}) and one on {\sc Wilkinson} ({\sc diffvec-wk}), omitting all scales where $a_{ext}$ or $a_{mild}$ are present in {\sc Crowd}. We do the same for the other datasets.

To obtain the $\overrightarrow{dVec}$ of a $s$ with static  embeddings, we simply calculate the difference between the word2vec embeddings of $a_{ext}$ and $a_{mild}$ in $s$.

\paragraph{Results} For evaluation, we use the same metrics as in Section \ref{subsec:fix_boundaries}. We  compare our results to the {\sc freq} and {\sc sense} baselines,  
 and 
to the best results obtained by \citet{cocos-etal-2018-learning}  who use information obtained from lexico-syntactic patterns, a lexicon annotated with intensity (SO-CAL) \citep{taboada2011lexicon}, and paraphrases from PPDB.\footnote{We do not report Spearman's $\rho$ from \citet{cocos-etal-2018-learning} because it was calculated differently: They measure it a single time for each dataset, treating each adjective as a single data point.} Results are presented in Table \ref{tab:diffvec_results}.
The {\sc diffvec} method gets 
remarkably high performance compared to previous results, 
especially when 
$\overrightarrow{dVec}$ is calculated 
with BERT embeddings. With the exception of Kendall's $\tau$ and pairwise accuracy on the {\sc deMelo} dataset, {\sc diffvec} 
outperforms results from previous work and the baselines across the board. 
We believe the lower correlation scores on the {\sc deMelo} dataset to be  
 due to the large amount of ties present in this 
dataset: 
44\% of scales in {\sc deMelo} contain ties, versus 30\% in {\sc Crowd} and 0\% in {\sc Wilkinson}, where we obtain better results. 
Our models cannot easily predict ties using similarities which are continuous values. 
To 
check whether our assumption is correct, we make a simple adjustment to {\sc diffvec} so that it can propose ties 
if the vectors of two adjectives are similarly close to 
$\overrightarrow{dVec}$.
Overall, this results in a small decrease in pairwise accuracy and a slight 
increase in correlation in {\sc deMelo} and {\sc Crowd}. 
Complete results of this additional evaluation are given 
in Appendix \ref{app:ties}. 

The 
composition of the {\sc sent-set}s  used for building BERT representations 
also plays a role on  model  performance. Overall, the selection method described in Section \ref{sec:ctxt_representations} offers a slight advantage over random selection, with  
ukWaC and Flickr sentences improving 
performance on different datasets. 
Note, however, that results for Flickr 
are calculated on the scales for which sentences were available (96.56\% of {\sc deMelo} scales and 86.08\% from {\sc Crowd}). 

The best-performing BERT layers are generally 
situated in the upper half of the Transformer network. The only exception is {\sc diffvec-wk} with the Flickr {\sc sent-set} on {\sc deMelo}, where all layers perform similarly. 
The 
{\sc freq} and {\sc sense} 
baselines  
get lower performance than our method with BERT embeddings. 
{\sc sense} 
manages to give results comparable  to  {\sc diffvec} with static embeddings 
and to previous work \citep{cocos-etal-2018-learning}  in one dataset ({\sc Crowd}),  but is still outperformed by {\sc diffvec} with contextualised representations.  

We can also compare our results to those obtained  by a purely pattern-based method on the same datasets, reported by \newcite{cocos-etal-2018-learning}. 
This method performs well on {\sc deMelo} ($\tau=$ 0.663) because of its high coverage on this dataset, which was 
compiled by finding adjective pairs that also match lexical patterns. 
The performance of the pattern-based method is much lower than that of our models in the other two datasets ($\tau=$ 0.203 on {\sc Crowd}, $\tau=$ 0.441 on {\sc Wilkinson}), and its coverage goes down to 11\% on {\sc Crowd}. This highlights the limitations of the approach, as well as the 
efficiency of our model which combines high performance and coverage.

\subsection{Further Exploration of {\sc diffvec}} \label{subsec:analysis}

Given the high performance of the {\sc diffvec} method in the ranking task, 
we  carry out additional experiments 
to  explore the impact that 
the choice of scales and sentences 
has on the intensity vector quality. 
We test the method with 
a $\overrightarrow{dVec}$ vector built from a single $a_{ext}-a_{mild}$ 
pair of either 
positive (\textit{awesome-good}) or negative (\textit{horrible}-\textit{bad}) polarity, 
that we respectively call {\sc diffvec-1 $(+)$/$(-)$}. 
 We also experiment with increasing the number of scales, 
adding 
\textit{ancient}-\textit{old}, 
\textit{gorgeous}-\textit{pretty} and \textit{hideous}-\textit{ugly} to form {\sc diffvec-5}. 
The  scales are from {\sc Wilkinson}, so we exclude this dataset from the evaluation. 

\begin{table}[t!]
    \centering
    \scalebox{0.90}{
     
      \begin{tabular}{ c c | c | ccc }
     & & \multicolumn{1}{c}{} & \multicolumn{3}{c}{{\sc deMelo}}  \\ 
     & & \# Scales & {\sc p-acc} & $\tau$ & $\rho_{avg}$ \\
     \hline
     \parbox[t]{3mm}{\multirow{9}{*}{\rotatebox[origin=c]{90}{BERT}}} &
    \parbox[t]{3mm}{\multirow{3}{*}{\rotatebox[origin=c]{90}{ukWaC}}}
    & 1 $(+)$  & 
     0.653$_{9}$ & 0.438$_{9}$ & 0.489$_{11}$ \\
    & & 1 $(-)$ &     0.611$_{10}$ & 0.350$_{10}$ & 0.424$_{11}$  \\
    & & 5 &  
    0.650$_{10}$ & 0.430$_{10}$ & 0.514$_{10}$ \\
      
    \cdashline{2-5}
    
    & \parbox[t]{3mm}{\multirow{3}{*}{\rotatebox[origin=c]{90}{Flickr}}}
& 1 $(+)$ &0.656$_{8}$ & 0.449$_{8}$  &  0.504$_{8}$ \\
    & & 1 $(-)$ & 0.600$_{3}$ & 0.324$_{3}$ & 0.375$_{5}$ \\
    & & 5 & 0.647$_{12}$ & 0.426$_{12}$ & 0.498$_{11}$ \\
    \cdashline{2-5}
     
    & \parbox[t]{3mm}{\multirow{3}{*}{\rotatebox[origin=c]{90}{Random}}}
& 1 $(+)$ & \textbf{0.659}$_{11}$ & \textbf{0.451}$_{11}$ & 0.493$_{11}$  \\
    & & 1 $(-)$ & 0.608$_{12}$ & 0.340$_{12}$ & 0.421$_{10}$  \\
    & & 5 &  0.653$_{11}$ &  0.442$_{11}$ &  \textbf{0.538}$_{10}$ \\

     \hline

    \parbox[t]{3mm}{\multirow{3}{*}{\rotatebox[origin=c]{90}{word2vec}}} & &
    1 $(+)$  & 0.602 & 0.334 & 0.364  \\
    & & 1 $(-)$ & 0.613 & 0.359 & 0.412  \\
    & & 5 & 0.641 & 0.415 & 0.438 \\

    \end{tabular}}
    
\vspace{0.5cm}
    \centering
    \scalebox{0.90}{
      \begin{tabular}{ c  c | c | ccc}
    & & \multicolumn{1}{c}{} & \multicolumn{3}{c}{{\sc Crowd }} \\ 
     
    & & \# Scales & {\sc p-acc} & $\tau$ & $\rho_{avg}$ \\
     \hline
    \parbox[t]{3mm}{\multirow{9}{*}{\rotatebox[origin=c]{90}{BERT}}}  &\parbox[t]{3mm}{\multirow{3}{*}{\rotatebox[origin=c]{90}{ukWaC}}}
    & 1 $(+)$ & \textbf{0.709}$_{12}$ & \textbf{0.611}$_{12}$ & \textbf{0.670}$_{12}$ \\
    & & 1 $(-)$ &  0.648$_{10}$ & 0.477 & 0.507$_{10}$  \\
    & & 5 &  0.700$_{11}$ & 0.595$_{10}$ & 0.673$_{10}$ \\
      
    \cdashline{2-5}
    
    & \parbox[t]{3mm}{\multirow{3}{*}{\rotatebox[origin=c]{90}{Flickr}}}
&  1 $(+)$ & 0.676$_{12}$ & 0.552$_{8}$ & 0.612$_{8}$ \\
    & & 1 $(-)$ & 0.641$_{9}$ & 0.470$_{9}$ & 0.502$_{9}$  \\
    & & 5 &  0.692$_{11}$ & 0.587$_{11}$ & 0.640$_{11}$ \\
    \cdashline{2-5}
     
     &  \parbox[t]{3mm}{\multirow{3}{*}{\rotatebox[origin=c]{90}{Random}}}
& 1 $(+)$ & 0.691$_{11}$ & 0.570$_{11}$ & 0.658$_{11}$ \\
    & & 1 $(-)$ & 0.655$_{10}$ & 0.490$_{10}$ & 0.514$_{12}$  \\
    & & 5 & 0.694$_{11}$ &  0.582$_{11}$ &  0.653$_{11}$ \\


    \hline
    
    \parbox[t]{3mm}{\multirow{3}{*}{\rotatebox[origin=c]{90}{word2vec}}} & &
    1 $(+)$  & 0.624 & 0.419  & 0.479  \\
    & & 1 $(-)$ & 0.661 & 0.506 & 0.559  \\
    & & 5 & 0.688 & 0.559 & 0.601

    \end{tabular}}
    \caption{Results of {\sc diffvec} on {\sc deMelo} and on {\sc Crowd} using a single positive (1 $(+)$) or negative (1 $(-)$) $a_{ext} - a_{mild}$ pair, and five pairs (5).}
    \label{tab:diffvecanalysis_results}
\end{table}

Results 
are given in Table \ref{tab:diffvecanalysis_results}. 
We observe that 
a  small number 
of 
word pairs is enough  to build a $\overrightarrow{dVec}$ 
with competitive performance. Interestingly, 
{\sc diffvec-1 $(+)$} with random sentences 
obtains the best pairwise accuracy on {\sc deMelo}. The fact that the method performs so well with just a few pairs 
(instead of a whole dataset as in Table \ref{tab:diffvec_results})  
is very encouraging, making our approach easily applicable to other datasets and languages.

A larger number of scales is 
beneficial for the method with static word2vec embeddings, which seem to better capture intensity on the negative scale. 
For BERT, instead, 
intensity modeled using a positive pair 
gives best results across the board. 
The use of 
five pairs of mixed polarity improves results over a single negative pair, and has comparable performance to the single positive one. 


Finally, we compare the performance of 
{\sc diffvec-1 $(+)$/$(-)$} and {\sc diffvec-5} when 
the contextualised representations are extracted from a single sentence instead of ten.  
Our main observation is that reducing the number of sentences 
harms performance, especially when 
the sentence used is randomly selected.
Detailed results are included in Appendix \ref{app:singlesentence}. 


\section{Indirect Question Answering} \label{sec:qa}

We conduct an additional evaluation in order to assess how useful {\sc diffvec} 
adjective rankings can be in a real application. As in \citet{cocos-etal-2018-learning}, we address Indirect Question Answering (QA)  \citep{demarneffe:10}. 
The task consists in  interpreting indirect answers to YES/NO questions involving  
scalar adjectives. 
These 
do not straightforwardly convey a YES or NO 
answer, but the intended reply can be inferred. For example, if someone is asked ``{\it Was it a good ad?}'' and replies ``{\it It was a great ad}'', the answer is YES. 
This makes Indirect QA a good fit for scalar adjective ranking evaluation since it allows to directly assess a model's capability to detect the difference in intensity and direction (positive or negative) in an adjective pair.

\begin{table}[t]
    \centering
    \scalebox{0.81}{
    \begin{tabular}{c c l | c ccc}
         & 
         & Method & Acc & P & R & F\\
         \hline
        \parbox[t]{2mm}{\multirow{12}{*}{\rotatebox[origin=c]{90}{BERT}}} & 
        \parbox[t]{2mm}{\multirow{4}{*}{\rotatebox[origin=c]{90}{ukWaC}}}
        & {\sc diffvec-1} $(+)_{10}$ & 0.715 & 0.677 & 0.692 & 0.685 \\
         & & {\sc diffvec-dm}$_{12}$ & 0.707  & 0.670 & 0.689 & 0.678 \\
         & & {\sc diffvec-cd}$_{12}$ & 0.675 & 0.635 & 0.648& 0.642 \\
         & & {\sc diffvec-wk}$_{11}$ & \textbf{0.740} & \textbf{0.712} & \textbf{0.739} & \textbf{0.725} \\ \cdashline{2-7}
        & 
         \parbox[t]{2mm}{\multirow{4}{*}{\rotatebox[origin=c]{90}{Flickr}}} & {\sc diffvec-1} $(+)_{9}$ & 0.699 & 0.663 & 0.680  & 0.672  \\ 
         & & {\sc diffvec-dm}$_{11}$ & 0.699 & 0.659 & 0.673 & 0.666 \\
         & & {\sc diffvec-cd}$_{10}$ & 0.691 & 0.653 & 0.667 & 0.660 \\
         & & {\sc diffvec-wk}$_{5}$ & 0.683 & 0.646 & 0.661 & 0.654 \\
         \cdashline{2-7}
        
         & 
         \parbox[t]{2mm}{\multirow{4}{*}{\rotatebox[origin=c]{90}{Random}}} 
         & {\sc diffvec-1} $(+)_{9}$ & 0.715 & 0.677 & 0.692 & 0.685 \\
         & & {\sc diffvec-dm}$_{10}$ & 0.724  & 0.691 & 0.713  & 0.702 \\
         & & {\sc diffvec-cd}$_{12}$ & 0.667 & 0.629 & 0.642 & 0.636 \\
         & & {\sc diffvec-wk}$_{11}$ & 0.699  & 0.667 & 0.688 & 0.677 \\
         \hline
        \parbox[t]{2mm}{\multirow{4}{*}{\rotatebox[origin=c]{90}{word2vec}}} & & {\sc diffvec-1} $(+)$ & 0.667 & 0.633 & 0.650 & 0.641 \\
        & &  {\sc diffvec-dm} & 0.602 & 0.554 & 0.559 & 0.557 \\
        &  &  {\sc diffvec-cd} & 0.593 & 0.548& 0.553  & 0.551 \\
        &  &  {\sc diffvec-wk} & 0.585 & 0.543 & 0.547 & 0.545 \\
         \hline
         \parbox[t]{2mm}{\multirow{6}{*}{\rotatebox[origin=c]{90}{Baselines}}}  & & {\sc freq} &  0.593  & 0.548  & 0.553 & 0.551 \\
         & & {\sc sense}  & 0.593 & 0.560 & 0.568 & 0.564  \\
         & & {\sc maj} & 0.691 & 0.346 & 0.500 & 0.409 \\
          & & \vspace{1mm} $Previous_1$ 
         & 0.610 & 0.597 & 0.594 & 0.596 \\
         & & \vspace{1mm} $Previous_2$ 
         &  0.728 & 0.698 & 0.714 & 0.706 \\
         & & 
         \vspace{1mm} $Previous_3$  & 0.642 & 0.710 & 0.683 & 0.684 \\
    \end{tabular}}
    \caption{Results of our {\sc diffvec} method with contextualised (BERT) and static (word2vec) embeddings on the indirect QA task. We compare to the frequency, polysemy and majority baselines, and to results from previous work. $Previous_1$ stands for \newcite{demarneffe:10}, $Previous_2$ for \newcite{kim-de-marneffe-2013-deriving} (the only result on 125 pairs), $Previous_3$ for \newcite{cocos-etal-2018-learning}.}
    \label{tab:qa_results}
\end{table}

We use the \citet{demarneffe:10} dataset for evaluation, which consists of 125 QA pairs manually annotated with their implied answers (YES or NO). 
We  adopt a  decision procedure similar to the one proposed by 
\citet{demarneffe:10}.
We compute the BERT embeddings of the adjective in the question ($a_{q}$) and the adjective in the answer ($a_{a}$).
If $a_{a}$ 
(e.g., \textit{great}) has the same or higher intensity than 
$a_{q}$ (e.g., \textit{good}) the prediction is YES; otherwise, the prediction is NO. If the answer contains a negation, we switch YES to NO, and NO to YES. In previous work, indirect QA evaluation was performed on 123 or 125 examples, depending on whether cases labelled as ``uncertain'' were included \cite{demarneffe:10,kim-de-marneffe-2013-deriving,cocos-etal-2018-learning}. We report all available results from previous work, and our 
scores on the 123 YES/NO examples 
as in the most 
recent work by 
\citet{cocos-etal-2018-learning}. 
We report results using {\sc diffvec} with the adjustment for ties, where two adjectives are considered to be of the same intensity if they are similarly close to $\overrightarrow{dVec}$ ($diffsim$ = sim($\overrightarrow{dVec}$, $\overrightarrow{a_{q}}$) $-$ sim($\overrightarrow{dVec}$, $\overrightarrow{a_{a}}$)). If the absolute value of $diffsim <$ 0.01, we count them as a tie.
We compare our method to previous results, 
to {\sc freq} and {\sc sense}, and to a baseline predicting always the majority label (YES). 
Results of this evaluation are given in Table \ref{tab:qa_results}. {\sc diffvec} with BERT embeddings outperforms 
the baselines and all previous approaches, and presents a clear advantage over {\sc diffvec} with static word2vec representations. Best performance is obtained  when $\overrightarrow{dVec}$ is obtained from the Wilkinson dataset ({\sc diffvec-wk}). The $\overrightarrow{dVec}$ obtained from {\sc Crowd} seems to be of lower quality. 
{\sc diffvec-cd} and {\sc diffvec-dm} improve over the baselines 
but do not achieve higher performance than the model of  
\citet{kim-de-marneffe-2013-deriving}.

\section{Discussion}


Our initial exploration of the  knowledge encoded in BERT representations about scalar adjectives using {\sc Bertsim} (Section \ref{subsec:fix_boundaries}) showed they can successfully rank them by intensity. 
Then our {\sc diffvec} method (Sections \ref{subsec:ranking_noboundaries} and \ref{subsec:analysis}) outperformed {\sc Bertsim}, providing even better ranking predictions with as few resources as a single adjective pair. This difference can be due to the composition of the vectors in the two cases.  The $a_{ext}$ representation in {\sc Bertsim} contains information about the meaning of the extreme adjective alongside its intensity, while the $\overrightarrow{dVec}$ vector is a cleaner representation of intensity: The 
 subtraction of 
$\overrightarrow{a_{mild}}$ from 
$\overrightarrow{a_{ext}}$ 
removes the common core meaning expressed by their scale (e.g., {\sc beauty}, {\sc temperature}, {\sc size}). 
Consequently, $\overrightarrow{dVec}$ is a pure and general representation of intensity which can successfully serve to 
rank adjectives from any scale, as shown by our results. 
The {\sc diffvec} method 
can estimate adjectives' relative intensity 
on the fly, and performs better than the {\sc Bertsim} model which needs a reference point to propose a ranking. 
It does not use any external knowledge source -- a requirement in 
previous approaches -- and one of its highest performing variations ({\sc diffvec-1} $(+)$) 
makes best 
quality predictions 
with a single  adjective pair example.

Our 
assumption 
concerning the need for the sentences used for extracting BERT representations 
to be a good semantic fit for adjectives in a scale, has not been confirmed by our evaluation. 
Precisely, differences 
between our methods when relying on carefully vs randomly selected sentences are minor. 
This might be due to several reasons: One is 
that although BERT representations are contextualised, they also encode 
knowledge about the meaning and intensity of 
words acquired through pre-training, 
independent of the new context of use. 
Another possible explanation is that due to the skewed distribution of word senses  \citep{Kilgarriff2004,mccarthy-etal-2004-finding}, a high proportion of our randomly selected sentences 
might contain instances of the adjectives 
in their most frequent sense. If this is also the  meaning of the corresponding scale, then the  sentences are a good fit. 


The {\sc diffvec-1} $(+)$ method,  which uses a vector derived from a single positive pair, yields consistently better results than {\sc diffvec-1} $(-)$ which relies on a single negative pair. 
To better understand this difference in performance, we 
examine the composition of {\sc deMelo} and {\sc Crowd}, specifically whether there is an imbalance in terms of polarity  as reflected in the frequency of positive vs negative adjectives in the two datasets.  
We check the polarity of the adjectives 
in two sentiment lexicons: SO-CAL \cite{taboada2011lexicon} and AFINN-165 \cite{nielsen2011new}. 
The two lexicons cover a portion of 
the adjectives in {\sc deMelo} and {\sc Crowd}: 68\% and 79\%, respectively. The {\sc deMelo} dataset is well-balanced 
in terms 
of positive 
and negative adjectives: 51\% and 49\% of the covered adjectives fall in each category. 
In {\sc Crowd}, we observe a slight skew towards positive: 61\% vs 39\%. 
According to this analysis, the difference in performance between the two methods could only partially be explained by an imbalance in terms of polarity.

We perform an additional analysis based on the Google Ngram frequency of the positive and negative words that were used for deriving {\sc diffvec}. 
The adjectives {\it good} (276M) and \textit{awesome} (10M) are more frequent than 
\textit{bad} (65M) and \textit{horrible} (4M).
In fact, we find that the 1,000 most frequent positive words in SO-CAL and AFINN are, on average, much more frequent (18M) than the 1,000 most frequent negative words (8M). 
Word frequency has a direct impact on word representations, 
since having access to sparse information about a word's usages 
does not allow the model to acquire 
rich information about its linguistic properties as in the case of frequent words. 
The high frequency of \textit{good} and \textit{awesome} results in better quality representations than the ones obtained for their antonyms, and could explain to some extent the improved performance of 
{\sc diffvec}-1 $(+)$ compared to {\sc diffvec}-1 $(-)$ with BERT embeddings. However, this analysis does not explain the difference in the performance of {\sc diffvec} $(+)$ and $(-)$ between BERT and word2vec. This would require a better understanding of how words with different polarity (antonyms) are represented in BERT's space compared to word2vec, and how negation affects their representations. We leave these explorations for future work.

\begin{figure}
    \centering
    \includegraphics[width=\columnwidth]{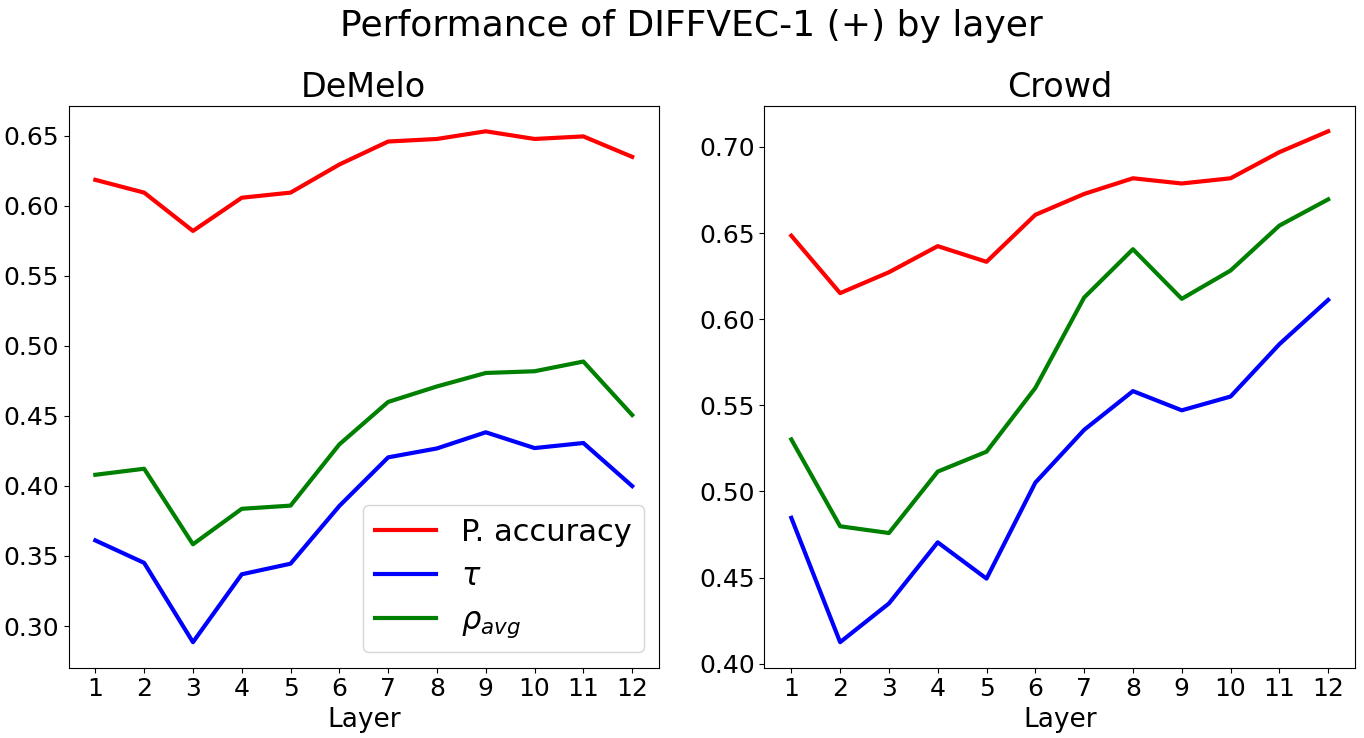}
    \caption{Performance of {\sc diffvec-1 $(+)$} with ukWaC sentences across BERT layers.}
    \label{fig:bylayer_diffvec}
\end{figure}

Regarding the performance of different BERT layers, 
we observe that 
knowledge relevant for scalar adjective ranking 
is situated in 
the last layers of the Transformer network. Figure \ref{fig:bylayer_diffvec} shows how the performance of {\sc diffvec-1 $(+)$} changes across  
different BERT layers: model predictions improve after 
layer 3, and performance 
peaks in one of the last four layers. This is in accordance with the findings of  \newcite{tenney-etal-2019-bert} 
that semantic information is mainly located in the upper layers of the model, but is more spread across the network than syntactic information which is contained in a few middle layers.



\section{Conclusion}

We have shown that BERT representations encode rich information about
the intensity of scalar adjectives which can be efficiently used for their ranking. Although our method is simple and resource-light, solely relying on an intensity vector which can be derived from as few as a single example, 
it clearly outperforms previous work on the scalar adjective ranking and  Indirect Question Answering tasks. 
Our performance analysis across BERT layers highlights that 
the lexical semantic knowledge needed  
for these tasks is mostly located in the higher layers of the BERT model. 

In future work, we plan to 
extend our methodology to 
new languages, and experiment with multilingual and language specific BERT models.  
To create scalar adjective resources in new languages, we could either translate the English datasets or mine
adjective scales from starred product reviews as in \citet{demarneffe:10}.
Our intention is also to  
address adjective ranking in full scales (instead of half-scales) 
and evaluate the capability of contextualised representations to detect polarity.


\newpage 
\section*{Acknowledgements}


\setlength\intextsep{0mm}

\begin{wrapfigure}[]{l}{0pt}
\includegraphics[scale=0.3]{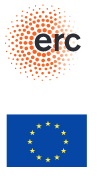}
\end{wrapfigure}

This work has been supported by the French National Research Agency under project ANR-16-CE33-0013. The work is also part of the FoTran project, funded by the European Research Council (ERC) under the European Union’s Horizon 2020 research and innovation programme (grant agreement \textnumero ~771113). We thank the reviewers for their thoughtful comments and valuable suggestions.

\bibliography{emnlp2020,anthology}
\bibliographystyle{acl_natbib}

\appendix

\section{Hearst Patterns} \label{app:hearst}

\begin{figure}[t!]
    \centering
    
    \includegraphics[height=17cm]{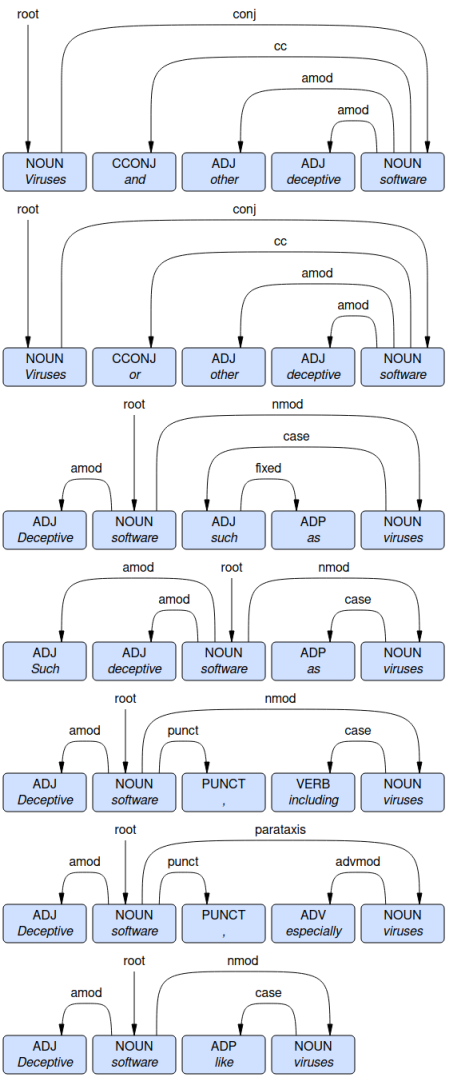}
    \caption{Dependency structure of Hearst patterns.}
    \label{fig:dependencies}
\end{figure}

Figure \ref{fig:dependencies}  
illustrates the dependency structure 
of the following Hearst patterns: 

\begin{itemize}
    \item \textit{[NP] and other [NP]}
    \item \textit{[NP] or other [NP]}
    \item \textit{[NP] such as [NP]}
    \item \textit{Such [NP] as [NP]}
    \item \textit{[NP], including [NP]}
    \item \textit{[NP], especially [NP]}
    \item \textit{[NP] like [NP]}

\end{itemize}

\noindent We use these patterns to detect sentences where adjective substitution should not take place, as described in Section \ref{sec:sentence_cleaning} of the paper. We remove these sentences from our ukWaC and Flickr datasets.\footnote{Graphs in Figure \ref{fig:dependencies} were created with the visualisation tool available at \url{https://urd2.let.rug.nl/~kleiweg/conllu/}}

\section{Evaluation of Sentence Selection Methods} \label{app:evalsentence}

To identify the most appropriate method for selecting 
sentences where all adjectives in a scale fit, we use data from the Concepts in Context (CoInCo) corpus \cite{kremer2014substitutes}. CoInCo contains sentences where  
content words have been 
manually annotated with substitutes 
which come with a frequency score indicating the number of annotators who proposed each substitute. We collect instances of adjectives, nouns and verbs in their base form.\footnote{This filtering serves to control for morphological variation which could result in unnatural substitutions since CoInCo substitutes are in lemma form.} For a word $w$, we form 
instance pairs ($w_{i}$-$w_{j}$ with $i \neq j$) with similar meaning as reflected in their shared substitutes. We allow for up to two unique substitutes per instance, which we assign to the other instance in the pair with zero frequency. We keep instances with $n$ substitutes, where  2 $\leq$ $n$ $\leq$ 8 (the lowest and highest number of adjectives in a scale). This results in  5,954 pairs.

We measure the variation in an instance pair in terms of substitutes using the \textit{coefficient of variation} ({\sc var}). {\sc var} is the ratio of the standard deviation to the mean and 
is, therefore, independent from  the unit used. A higher {\sc var}  indicates that not all substitutes are good choices in a context. We keep the 500  pairs with the highest {\sc var} difference, where one sentence is a better fit for all substitutes than the other. For example, \textit{private}, \textit{individual} and \textit{person} were proposed as substitutes for {\it personal}  in 
``\textit{personal insurance lines}'', but \textit{private} was the preferred choice for
``\textit{personal reasons}''. 
The tested methods must identify which sentence in a pair is a better fit for all substitutes.

For sentence selection, we experiment with the three fluency calculation  methods presented 
in Section \ref{sec:sentence_cleaning}: 
{\sc BERTprob} (the BERT probability of each substitute to be used in the place of the [MASK] token); {\sc BERTppx} (the perplexity assigned by BERT to the sentence generated through substitution); and {\sc context2vec} (the cosine similarity between the context2vec representations of a substitute and the context).

We also test {\sc var} and standard deviation ({\sc std}) as metrics for measuring variation in the fluency scores assigned to a sentence pair by the three methods.
We evaluate the sentence selection methods and variation metrics on the 500 pairs 
retained from CoInCo. We report their accuracy, calculated as the proportion of pairs where a method correctly guesses the instance in a pair with the lowest variation.
We compare results to those of a baseline that always proposes the first instance in a pair.
The results in Table \ref{tab:coinco_results} show 
that the task is difficult for all 
methods. Their accuracy is slightly higher than the baseline accuracy, 
which 
outperforms  {\sc BERTprob} with {\sc var}. The 
combination that gives 
best accuracy is {\sc context2vec} 
with {\sc std} 
(0.594). We use this combination of metrics in our experiments.

\begin{table}
    \centering
    \scalebox{0.9}{
    \begin{tabular}{lc|c}
    Method & Variation Metric & Accuracy \\
    \toprule
         \parbox{2cm}{\vspace{2mm}\multirow{2}{0pt}{\sc{BERTprob}}} & {\sc std} & 0.524 \\
          & {\sc var} & 0.488 \\
         \hline
         \parbox{2cm}{\vspace{2mm}\multirow{2}{0pt}{{\sc BERTppx}}} & {\sc std} & 0.518 \\
         & {\sc var} & 0.536 \\
         \hline
         \parbox{2cm}{\vspace{2mm}\multirow{2}{0pt}{{\sc context2vec}}} & {\sc std} & \textbf{0.594} \\
         & {\sc var} & 0.588 \\
         \hline
         1st sentence Baseline &  & 0.506 
    \end{tabular}
    }
    \caption{Accuracy of the three fluency calculation methods on the 500 sentence pairs collected from CoInCo. Comparison to a first sentence baseline.}
    \label{tab:coinco_results}
\end{table}

\begin{table*}[t]
    \centering
    \scalebox{0.85}{
    \begin{tabular}[width=\textwidth]{p{3.7cm}p{1.1cm}p{13cm}}
    \multicolumn{3}{l}{{\bf Scale: } \textit{wrong} $\rightarrow$ \textit{immoral} $\rightarrow$ \textit{sinful} $\rightarrow$ \textit{evil} } \\
    \hline
    Method & Corpus & Sentences \\
    \hline
    \multirow{2}{0pt}{\parbox{3.7cm}{context2vec-{\sc std}}} & ukWaC & I believe that war is \textit{immoral}. \\
    & Flickr & This boy was on the \textit{wrong} end of this snowball fight. \\ \cdashline{1-3}
    
    Random & ukWaC & The author saw him and let him thru but not his mate as he had queued the \textit{wrong} way. \\
    \toprule
    \multicolumn{3}{l}{{\bf Scale: } \textit{old} $\rightarrow$ \textit{obsolete} $\vert\vert$ \textit{outdated}} \\
    \hline
        Method & Corpus & Sentences \\
    \hline
   \multirow{3}{0pt}{\parbox{3.7cm}{context2vec-{\sc std}}} & ukWaC & (...) Chekhov was misunderstood and frequently seen by critics as merely an irreverent recorder of an \textit{obsolete} way of life (...) \\
    & Flickr & Two preschool aged boys are looking at an \textit{old} locomotive.\\ \cdashline{1-3}
    Random & ukWaC & (...) rustic dialogue and good \textit{old} fashioned laughter (...)\\
    
    \end{tabular}}
    \caption{Examples of sentences from our {\sc sent-set}s selected with the context2vec-{\sc std} method compared to sentences randomly selected from ukWaC.}
    \label{tab:example_sentences}
\end{table*}

\begin{table*}[h!]
    \centering
    \scalebox{0.9}{
     \scalebox{0.9}{
      \begin{tabular}{c c | c | ccc | ccc | ccc}
     & & \multicolumn{1}{c}{} & \multicolumn{3}{c}{{\sc deMelo (dm)}} & \multicolumn{3}{c}{{\sc Crowd (cd)}} & \multicolumn{3}{c}{{\sc Wilkinson (wk)}} \\ \cline{1-12}
     & 
     & Method & {\sc p-acc} & $\tau$ & $\rho_{avg}$ & {\sc p-acc} & $\tau$  & $\rho_{avg}$ & {\sc p-acc} & $\tau$ & $\rho_{avg}$\\
     \hline
    \parbox[t]{2mm}{\multirow{9}{*}{\rotatebox[origin=c]{90}{BERT}}} & 
    \parbox[t]{2mm}{\multirow{3}{*}{\rotatebox[origin=c]{90}{ukWaC}}}
    & {\sc diffvec-dm} & 
     - & - & - & 
     0.733$_{8}$ & 0.673$_{8}$ & \textbf{0.749}$_{12}$ &
     0.885$_{6}$& 0.830$_{11}$ & 0.826$_{6}$ \\
     & & {\sc diffvec-cd} & 
     \textbf{0.644}$_{8}$ & \textbf{0.452}$_{8}$ & \textbf{0.518}$_{8}$ & 
     - & - & - & 
     0.820$_{10}$ & 0.721$_{11}$ & 0.780$_{11}$\\
     & & {\sc diffvec-wk} & 
     0.546$_{6}$ & 0.295$_{6}$ & 0.324$_{6}$ & 
     0.721$_{7}$ & 0.627$_{10}$ & 0.698$_{10}$ &
     - & - & - \\
    \cdashline{2-12}
    & \parbox[t]{2mm}{\multirow{3}{*}{\rotatebox[origin=c]{90}{Flickr}}} & {\sc diffvec-dm} & 
     - & - & - & 
     0.746$_{12}$ & \textbf{0.685}$_{12}$ & 0.718$_{8}$ &
     \textbf{0.902}$_{9}$ & \textbf{0.851}$_{9}$ & \textbf{0.871}$_{4}$ \\
     & &  {\sc diffvec-cd} & 
     0.605$_{11}$ & 0.388$_{11}$ & 0.465$_{11}$ & 
     - & - & - & 
     0.836$_{8}$ & 0.746$_{7}$ & 0.762$_{7}$ \\
    &  & {\sc diffvec-wk} & 
     0.541$_{2}$ & 0.296$_{1}$ & 0.299$_{1}$ & 
    0.702$_{8}$ & 0.647$_{8}$  & 0.710$_{8}$ &
    - & - & - \\
    \cdashline{2-12}
    & \parbox[t]{2mm}{\multirow{3}{*}{\rotatebox[origin=c]{90}{Random}}}
    & {\sc diffvec-dm} & 
     - & - & - & 
     \textbf{0.724}$_{9}$ & 0.652$_{9}$ & 0.719$_{8}$ &
     0.885$_{11}$ & 0.818$_{6}$ & 0.833$_{10}$ \\
     & &  {\sc diffvec-cd} & 
     0.619$_{8}$ & 0.412$_{8}$ & 0.488$_{8}$ & 
     - & - & - & 
     0.819$_{12}$ & 0.765$_{10}$ & 0.833$_{10}$ \\
     & & {\sc diffvec-wk} & 
     0.522$_{2}$ & 0.251$_{6}$ & 0.285$_{6}$ & 
     0.712$_{10}$ & 0.614$_{9}$ & 0.680$_{9}$ &
     - & - & - \\
     \hline
    \parbox[t]{2mm}{\multirow{3}{*}{\rotatebox[origin=c]{90}{\small word2vec}}} &  & {\sc diffvec-dm} & 
     - & - & - & 
     0.648 & 0.508 & 0.550 &
     0.754 & 0.583  & 0.655 \\
     & &  {\sc diffvec-cd} & 
     0.604 & 0.403 & 0.446 & 
     - & - & -  & 
     0.803 & 0.656  & 0.661 \\
     & & 
     {\sc diffvec-wk} & 
     0.568 & 0.329 & 0.402 & 
     0.606 & 0.414 & 0.445 &
     - & -  & - \\
     \hline
   
         
    \end{tabular}}}
    \caption{Results of our {\sc diffvec} adjective ranking method 
    on the  {\sc deMelo}, {\sc Crowd} and {\sc Wilkinson} datasets with the adjustment for ties. 
    We report results with contextualised (BERT) representations obtained from  different {\sc sent-set}s (ukWaC, Flickr, Random) and with static (word2vec) vectors.  
   }
    \label{tab:diffvec_results_tie_adjusted}
\end{table*}

Table \ref{tab:example_sentences} shows examples of sentences retained after this filtering for two adjective scales. 
{\sc context2vec} tends to favour sentences where all adjectives in a scale fit well. We also give an example of a sentence randomly selected from ukWaC (Random) for a  scale.  These sentences 
usually reflect a frequent sense of a word 
in the scale. 


\section{Adjustment for Ties} \label{app:ties}

Table \ref{tab:diffvec_results_tie_adjusted} contains results of the {\sc diffvec} method with the adjustment for ties. For two adjacent adjectives ($a_{i}$, $a_{j}$) in the ranking  
proposed by {\sc diffvec}, we check if their cosine similarities to $\overrightarrow{dVec}$ are very close ($diffsim$ = sim($\overrightarrow{dVec}$, $\overrightarrow{a_{i}}$) - sim($\overrightarrow{dVec}$, $\overrightarrow{a_{j}}$). If the absolute value of $diffsim <$ 0.01, 
we count them as a tie, meaning that 
$a_{i}$ and  $a_{j}$ are 
considered to be 
situated at the same intensity level. Note that this procedure may give different results when 
the pairwise comparison starts at different ends of the proposed ranking. 
We 
establish ties starting from the $a$ with lowest intensity in the ranking 
proposed by {\sc diffvec}.



\section{{\sc diffvec} with a Single Sentence} \label{app:singlesentence}

Table \ref{tab:diffvecanalysis_results_onesentence} contains results for 
{\sc diffvec-1 $(+)$/$(-)$} and {\sc diffvec-5} when using a single sentence for building $\overrightarrow{dVec}$. 

\begin{table*}[h!]
    \centering
    \scalebox{0.9}{
      \begin{tabular}{c  c | c | ccc | ccc }
     & \multicolumn{1}{c}{} & & \multicolumn{3}{c}{{\sc deMelo}} & \multicolumn{3}{c}{{\sc Crowd }} \\ \cline{1-9}
     & 
     & \# Scales & {\sc p-acc} & $\tau$ & $\rho_{avg}$ & {\sc p-acc} & $\tau$  & $\rho_{avg}$ \\
     \hline
    
     \parbox[t]{5mm}{\multirow{9}{*}{\rotatebox[origin=c]{90}{BERT}}} &
    \parbox[t]{3mm}{\multirow{3}{*}{\rotatebox[origin=c]{90}{ukWaC}}}
& 1 $(+)$ &  0.651$_{10}$ &  0.433$_{10}$ &  0.501$_{10}$ &  0.682$_{10}$ &  0.553$_{10}$ &  0.622$_{7}$\\
    & & 1 $(-)$ & 0.597$_{1}$ & 0.315$_{1}$ & 0.352$_{1} $ & 0.639$_{12}$ & 0.458$_{12}$ & 0.543$_{12}$  \\
    & & 5 & \textbf{0.655}$_{7}$ & \textbf{0.443}$_{7}$ & \textbf{0.530}$_{7}$ & \textbf{0.691}$_{11}$ & \textbf{0.575}$_{11}$ & \textbf{0.675}$_{11}$ \\
      
    \cdashline{3-9}
    &     \parbox[t]{3mm}{\multirow{3}{*}{\rotatebox[origin=c]{90}{Flickr}}}
& 1 $(+)$ & 0.639$_{9}$ & 0.410$_{9}$ & 0.432$_{9}$ & 0.676$_{8}$ & 0.550$_{8}$ & 0.604$_{8}$\\
    & & 1 $(-)$ & 0.602$_{3}$ & 0.329$_{3}$ & 0.372$_{3}$ & 0.629$_{4}$ & 0.443$_{4}$ & 0.479$_{4}$  \\
    & & 5 & 0.624$_{11}$ & 0.380$_{11}$ & 0.452$_{11}$ & 0.683$_{11}$ & 0.562$_{11}$ & 0.606$_{12}$ \\
    \cdashline{3-9}
    & 
     \parbox[t]{3mm}{\multirow{3}{*}{\rotatebox[origin=c]{90}{Random}}}
& 1 $(+)$ &  0.631$_{11}$ &  0.401$_{11}$ &  0.451$_{11}$ & 0.676$_{8}$ &  0.536$_{8}$ &  0.589$_{8}$ \\
    &  & 1 $(-)$ & 0.611$_{9}$ & 0.356$_{9}$ & 0.444$_{9}$ & 0.648$_{11}$ & 0.479$_{11}$ & 0.500$_{11}$  \\
    &  & 5 & 0.622$_{4}$ & 0.371$_{4}$ & 0.417$_{3}$ & 0.685$_{7}$ & 0.559$_{7}$ & 0.588$_{7}$ \\
     \cline{2-9}

    \hline
    \parbox[t]{3mm}{\multirow{3}{*}{\rotatebox[origin=c]{90}{word2vec}}} & & 
    1 $(+)$  & 0.602 & 0.334 & 0.364 & 0.624 & 0.419  & 0.479  \\
    & & 1 $(-)$ & 0.613 & 0.359 & 0.412 & 0.661 & 0.506 & 0.559  \\
    & & 5 & 0.641 & 0.415 & 0.438 & 0.688 & 0.559 & 0.601

    \end{tabular}}
    \caption{Results of {\sc diffvec} using a single positive  (1 $(+)$) or negative (1 $(-)$) adjective pair, and five pairs (5). These are results obtained with a  $\overrightarrow{dVec}$  built from only one sentence (instead of ten in Table \ref{tab:diffvecanalysis_results} of the paper).}
    \label{tab:diffvecanalysis_results_onesentence}
\end{table*}

\end{document}